\documentclass[journal]{IEEEtran}
\usepackage{cite}
\usepackage{epsfig}

\usepackage{amsmath}
\usepackage{amssymb}
\usepackage{color}
\usepackage{url}
\usepackage{makecell}
\usepackage{verbatim}
\usepackage{multirow}
\usepackage{tabularx}
\usepackage{subfigure}

\usepackage[colorlinks,linkcolor=red]{hyperref}

\usepackage{autobreak}
\usepackage{breqn}

\usepackage{booktabs} % 

\usepackage{soul}
\soulregister{\cite}7 %  
\soulregister{\citep}7 %  
\soulregister{\citet}7 %  
\soulregister{\ref}7 %  
\soulregister{\pageref}7 % 

\usepackage[table,x11names]{xcolor}

\ifCLASSINFOpdf
\else
\fi
\begin{document}

\title{High Efficiency Image Compression for Large Visual-Language Models}

\author{Binzhe Li, 
        Shurun Wang,
        Shiqi Wang,~\IEEEmembership{Senior Member,~IEEE}, 
        and Yan Ye,~\IEEEmembership{Senior Member,~IEEE} % <-this % stops a space

\thanks{Binzhe Li, and Shiqi Wang are with the Department of Computer Science, City University of Hong Kong, Hong Kong (e-mail: binzheli2-c@my.cityu.edu.hk; shiqwang@cityu.edu.hk).}
\thanks{Shurun Wang is  with Alibaba Group, 
Beijing, China (e-mail: shurun.wsr@alibaba-inc.com ).}
\thanks{Yan Ye is with Alibaba Group U.S., Sunnyvale, CA 94085 USA (e-mail: yan.ye@alibaba-inc.com).}  
}

\markboth{submitted to IEEE Transactions on Circuits and Systems for Video Technology}{Right}

\newcommand{\bl}[1]{{\color{blue}#1}}
\newcommand{\bbl}[1]{{\color{red}#1}}

\maketitle

\begin{abstract}
In recent years, large visual language models (LVLMs) have shown impressive performance and promising generalization capability in multi-modal tasks, thus replacing humans as receivers of visual information in various application scenarios. In this paper, we pioneer to propose a variable bitrate image compression framework consisting of a pre-editing module and an end-to-end codec to achieve promising rate-accuracy performance for different LVLMs. In particular, instead of optimizing an adaptive pre-editing network towards a particular task or several representative tasks, we propose a new optimization strategy tailored for LVLMs, which is designed based on the representation and discrimination capability with token-level distortion and rank.  
%to process the input image based on semantic information at the pixel level. 
%Moreover, a variable bitrate end-to-end image codec is developed to compress the preprocessed images. 
The pre-editing module and the variable bitrate end-to-end image codec are jointly trained by the losses based on semantic tokens of the large model, which introduce enhanced generalization capability for various data and tasks. {Experimental results demonstrate that the proposed framework could efficiently achieve much better rate-accuracy performance compared to the state-of-the-art coding standard, Versatile Video Coding.} Meanwhile, experiments with multi-modal tasks have revealed the robustness and generalization capability of the proposed framework.

\end{abstract}


\IEEEpeerreviewmaketitle

\section{Introduction}

\IEEEPARstart{L}{arge} visual-language models (LVLMs) have shown impressive success in a variety of multi-modal application domains. Images, which are typically featured with high data volume, are typically compressed for transmission before feeding to the LVLMs at the cloud end. 
Instead of supporting only a single task, LVLMs typically support multi-tasks simultaneously, which brings unprecedented challenges to image coding for machines~\cite{wang2024overview}. 
%These learning-based downstream applications further promote signal compression research that focuses on optimizing the accuracy of machine analytics tasks and the representation cost of visual signals.

In the past decades, as the default visual data communication solutions, existing image and video standards have been developed and facilitated to improve rate-distortion (RD) performance, such as H.264/AVC~\cite{wiegand2003overview}, H.265/HEVC~\cite{sullivan2012overview}, H.266/VVC~\cite{VVC}, and AVS~\cite{ma2022evolutionAVS}. 
Inspired by the rapid development of deep neural networks, many learning-based image and video codecs are proposed~\cite{lu2019dvc, MLVC, lin2022dmvc,guo2023enhanced,shi2022alphavc}, which have achieved comparable and even better RD performance compared with VVC~\cite{duan2023learned,meng2023learned}. 
Treating the human visual system (HVS) as the consumer of the visual signals is the default assumption for most traditional and learning-based codecs. However, when the receiver is replaced by machine vision, these codecs usually struggle to obtain satisfactory results. 
One apparent reason is that machine tasks typically require different semantic features, such that features for signal reconstruction toward human vision are not dedicated to machine vision. 
Therefore, a robust machine-oriented framework for efficiently compressing visual information is in a high demand, to adapt to the growing trend of employing a variety of artificial intelligence (AI) analytical models as the ultimate consumers of data.

\begin{figure*}[t]
\centering
\includegraphics[width=0.9\textwidth]{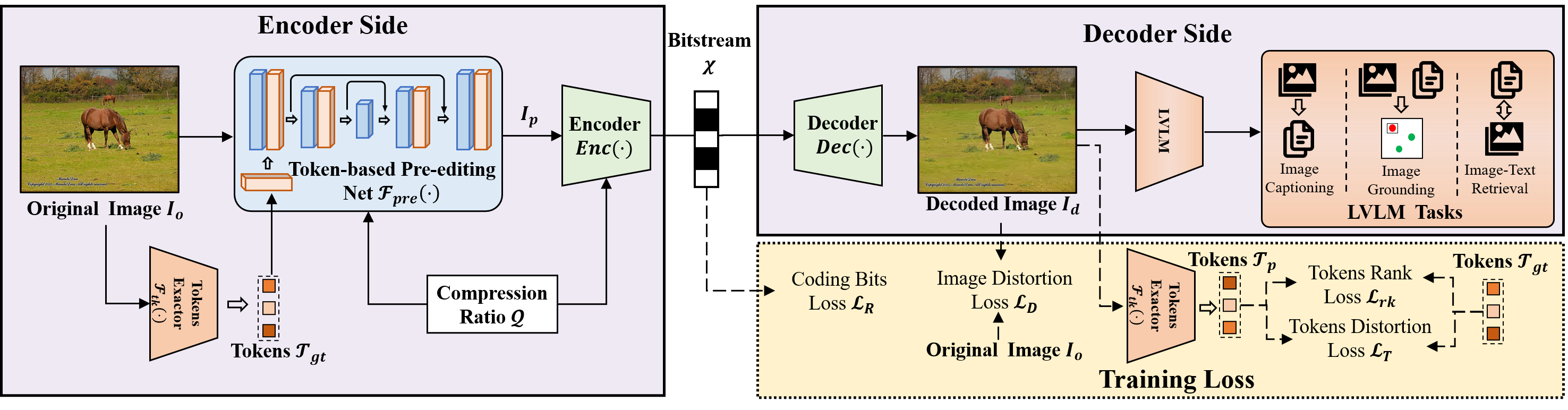}
\caption{ The paradigm of the proposed image compression scheme for LVLMs, including the encoder, decoder, and loss function design. {The pre-editing module includes the semantic tokens extractor and the pre-editing network, and the end-to-end codec is composed of the encoder and decoder for compressing and reconstructing the preprocessed image. The LVLMs are regarded as the ultimate receivers of the reconstructed images. }  }
\label{fig_paradigm}
\end{figure*}

{To facilitate downstream machine tasks with limited bandwidth, various efficient compression methods have been proposed to improve coding performance.} These methods are mainly divided into three categories.
The first category of methods applies the image/video enhancement networks as the pre-processing or post-processing modules~\cite{lu2022preprocessing,yang2024task,ahonen2021learned} without modifying the existing codecs and downstream analysis tasks. These enhancement networks are designed to remove unnecessary background information to save bitrate consumption or enhance the semantic features to achieve better accuracy for machine tasks.
The second category involves improving traditional or end-to-end codecs to optimize bitrate allocation and semantic feature representation~\cite{luo2021rate, huang2021visual, chamain2021end, wang2021towards, fischer2021boosting, wang2022deep}, thereby enhancing machine analysis performance. The codec is designed to reconstruct the machine task-friendly visual signal at the decoder side. The visual signal preserves the task-relevant semantic features, resulting in better rate-accuracy (RA) performance.  
Regarding the third research direction, the codecs are designed to transmit and reconstruct the compact feature representation of the machine vision task instead of the visual signal as an input to the downstream task~\cite{chen2019lossy,chen2020data,choi2020back,xing2020binary,suzuki2020deep}. The compression of compact features reduces inter-feature redundancy and lowers bitrate consumption. However, reconstructed intermediate features are specific to a limited number of task networks, lacking the generalization capability.

Most of the current approaches presume  specific tasks as downstream tasks after receiving the reconstructed signals/features, such as image classification~\cite{ResNet}, object detection~\cite{Girshick_2015_ICCV}, instance segmentation~\cite{he2017mask}. However, with the development of transformer-based large models~\cite{zhou2020unified,wang2021simvlm,wang2023pangu,wang2022ofa,wang2023one,li2023blip}, multi-modal tasks completed with LVLMs have been widely studied and deployed. Regarding video coding for machines, when the machine transitions from handling specific tasks to analyzing scenes in a more general manner, the compression algorithms need to be redesigned. 
{Compared to specific tasks, multi-modal tasks are not limited to particular object categories but can concentrate on richer semantic information.} For example, in detection and segmentation tasks, the color of a target is not semantically significant. {By contrast, in visual grounding or image-text retrieval tasks, the color of the target may be present in the textual description and serves as a pivotal piece of task-relevant semantic information, significantly aiding in the alignment of visual and textual modalities. Therefore, it is crucial to develop a compression scheme to overcome the limitations of existing methods with enhanced generalization capability.}

In this paper, we propose a dedicated image codec for LVLMs, as shown in Fig.~\ref{fig_paradigm}. 
{The primary idea is developing a token-based pre-editing framework and training the preprocessing network and codec jointly by incorporating a semantically-oriented optimization function. As such, we can preserve and enhance generic semantic information throughout the compression process, leading to an efficient image compression scheme for LVLMs.}
% The main idea is to build an image compression framework, including pre-editor and codecs, to achieve promising rate accuracy performance. 
Extensive experimental results show that the proposed framework achieves efficient and robust compression performance on different multi-modal tasks with different LVLMs.
The main contributions of this paper are summarized as follows,
\begin{itemize}

\item We propose an image compression scheme for LVLMs, consisting of the semantically driven pre-editor and the codec. This unique scheme is tailored to the growing trend of increasingly powerful artificial intelligence, and the whole process is fully optimized by the semantic information instead of fitting a particular analytical task. Experimental results show that the proposed scheme achieves more than 50\% bitrate savings at the same accuracy level when applying LVLMs as the ultimate receiver, and stronger generalization capability is exhibited in a variety of applications. 

%pre-editing module aims to enhance the features of visual signals while removing the  signals to generate suitable visual content for compression. The proposed variable bitrate codec module is designed to encode visual signals at low bit rates by learning the semantic signal distribution, and it reconstructs signals for better performance of large model analysis.

\item  We develop the pre-editing module guided by the large model tokens. The proposed pre-editing network is designed to preserve critical semantic information to maintain the machine task performance while discarding semantically irrelevant information to minimize bitrate consumption. This is accomplished by leveraging semantic tokens at different spatial levels, such that the characteristics of LVLMs can be fully exploited.

\item {We enhance the semantic consistency within the compression process by imposing supervision with the rank of large model tokens.} The design of this loss function is based on the assumption that the rank of tokens reflects the semantic abundance. By attempting to maintain the rank of tokens in the reconstructed visual signals, the performance of machine tasks with LVLMs can be further optimized.

\end{itemize}

%The rest of the paper is organized as follows. In Section~\ref{section:RealtedWork}, we take a thorough review and summary of the relevant works. In Section~\ref{section:Framework}, we introduce the variable bitrate image compression framework for LVLMs. The proposed pre-editing network and variable bitrate end-to-end codec are introduced in Section~\ref{section:Editing} and Section~\ref{section:Codec}, including the motivations, design philosophy, and principles, respectively. In Section~\ref{section:Loss}, the loss functions are presented for training the proposed framework.
%In Section~\ref{section:Experiments}, these experiments for various multi-modal tasks based on LVLMs are conducted to demonstrate the proposed framework's generalization capability and efficiency.
%The paper is concluded in Section~\ref{section:Conclusion}.

\section{Related Works}
\label{section:RealtedWork}

\subsection{Visual Signal Compression}
Visual signal compression plays a fundamental role in various image/video based applications, typically targeting at better perception quality with lower representation expense. Motivated by this, the compression of visual signals has been elaborately investigated in terms of signal processing in the past decades, especially in various image/video compression standards.  With the unprecedented development of deep learning in recent years, the efficiency of visual signal compression has also been explored by means of a variety of deep learning techniques.

\textbf{Image/video compression standards.} Over the past decades, visual signal compression has been driven by the development of image/video compression standards. For image compression, JPEG~\cite{wallace1992jpeg} has been widely deployed since 1992. JPEG 2000~\cite{rabbani2002jpeg2000} was further standardized for higher compression efficiency. Motivated by the new requirements of image compression on websites and mobile devices, multiple image codecs were further developed, such as WebP~\cite{lian2012webp} and BPG~\cite{albalawi2015hardware}, achieving obvious compression efficiency improvement. For video compression standards, there is a continuous evolution from H.262/MPEG-2~\cite{recommendation1995generic},
H.264/MPEG-4 AVC~\cite{wiegand2003overview} to H.265/HEVC~\cite{sullivan2012overview}. Based on this,  H.266/VVC~\cite{bross2021developments} was further developed and published in 2020, which is the state-of-the-art video compression standard.

\textbf{Deep learning based visual signal compression.} Inspired by the development of deep learning algorithms and hardware, deep learning-based visual signal compression has achieved remarkable progress, leveraging the powerful representation capability of artificial neural networks. 
%\bl{The deep learning methods have been integrated into the hybrid coding framework to enhance rate-distortion performance. In details, intra/inter prediction~\cite{laude2016deep,zhao2018enhanced,zhu2019generative,lei2022deep}, interpolation filtering~\cite{yan2017convolutional,liu2018one}, and in-loop filtering~\cite{jia2019content,pan2020efficient} were all investigated and improved with neural network based techniques.}
%Besides, p
Pioneering works have been proposed to compress images with a recurrent neural network (RNN)~\cite{toderici2015variable}. A block-based convolutional neural network (CNN) image compression model~\cite{liu2018cnn} was further proposed, achieving superior performance with JPEG. Moreover, a series of end-to-end image compression codecs were proposed, including~\cite{balle2016end},~\cite{balle2018variational}, and~\cite{minnen2018joint}, which are featured with generalized divisive normalization (GDN)~\cite{balle2015density}, variational hyper-prior structure and joint autoregressive hierarchical prior, respectively. Subsequently, the image compression performance is further improved with a discretized Gaussian Mixture Likelihoods~\cite{cheng2020learned}, which could parameterize the latent code distribution and improve the accuracy of the entropy model, achieving comparable performance compared with the still picture profile of VVC. Meanwhile, noticeable progress has been made in end-to-end video compression. An end-to-end video compression framework~\cite{lu2019dvc} was first proposed and achieved better performance compared with H.264/AVC. A feature-space video compression framework~\cite{hu2021fvc} was further proposed, achieving higher compression efficiency compared with HEVC. It is worth mentioning that the neural video compression with diverse contexts~\cite{li2023neural} could achieve superior performance compared with VVC under low delay configuration.

\subsection{Visual Signal Compression for Machine Vision}

With the successes of AI-based visual signal analysis in various machine vision applications, there is a shift for the ultimate consumer of visual signals from human perception to machine vision. Targeting to improve the coding efficiency for machine vision, a variety of techniques have been proposed, which could be categorized into codec improvement, feature compression, and processing for machine vision, depending on the reconstructed information at the decoder side.

\textbf{Visual signal codec for machine vision.} 
Recently, obvious progress has been made in deep learning-based visual signal codec for machine vision. To be specific, regarding the neural network structure for machine vision, the compression efficiency for machine vision is improved by means of a latent space masking network (LSMnet)~\cite{fischer2021boosting} and spatial-channel attention-based variable bitrate structure~\cite{wang2022deep}. In order to further improve the machine vision compression performance, the machine-oriented loss function is also investigated by means of the combination of signal level distortion and task loss function directly~\cite{fischer2023saliency}. The visual feature map distortion was further proposed to replace the task loss function in~\cite{le2021image}, avoiding the dependency of machine task labels. Moreover, the optimal weight parameter of various distortions in the joint loss function was explored in~\cite{wang2021end} with an iterative optimization algorithm, achieving obvious performance improvement for machine vision.

\textbf{Feature compression.} 
Machine vision is operated upon visual features instead of textures. As such, a common paradigm is to extract and compress the visual features for transmission rather than compressing the textures. Generally speaking, the visual features are typically more compact than textures, facilitating the simultaneous large-scale visual information transmission in front-end devices.
% An early attempt at the standardization for visual feature descriptors is Multimedia Content Description Interface, MPEG-7~\cite{chang2001overview}, which was initiated in 1998 for multimedia characteristic compact description, such as color and shape. 
To improve the visual search accuracy and efficiency, two standards, Compact Descriptors for Visual Search (CDVS)~\cite{duan2015overview}, and Compact Descriptors for Video Analysis (CDVA)~\cite{duan2018compact}, were developed for image and video search respectively. In recent years, a number of machine vision algorithms have been proposed for various machine analysis tasks, such as object detection~\cite{he2017mask}, instance segmentation~\cite{he2017mask}, and object tracking~\cite{wang2020towards}. To facilitate the aforementioned machine vision tasks, various algorithms have been proposed for the compact representation of deep learning features~\cite{ding2020joint, wang2020end, yang2021towards, kim2023end}. Moreover, the compression of deep learning features for machine vision is in the standardization process for VCM in MPEG~\cite{cfp_fcvcm}. 

\textbf{Processing toward machine vision.} 
It is a straightforward pipeline to process the visual signal toward machine vision without modifying the existing codec. The coding efficiency of image/video coding standards is boosted with various pre-post processing techniques, including the pre-processing~\cite{lu2022preprocessing, jvet_ac_0086}, spatial resampling~\cite{mpeg143_64124}, temporal resampling~\cite{cfp_res_60775} and post-processing~\cite{jvet_ag0212}. Moreover, a task-switchable pre-processor was designed in~\cite{yang2024task} to preserve important semantic information based on the specific characteristics of various downstream tasks, which supports the demand for multi-task applications in the real world.

\begin{figure*}[t]
\centering
\includegraphics[width=0.90\textwidth]{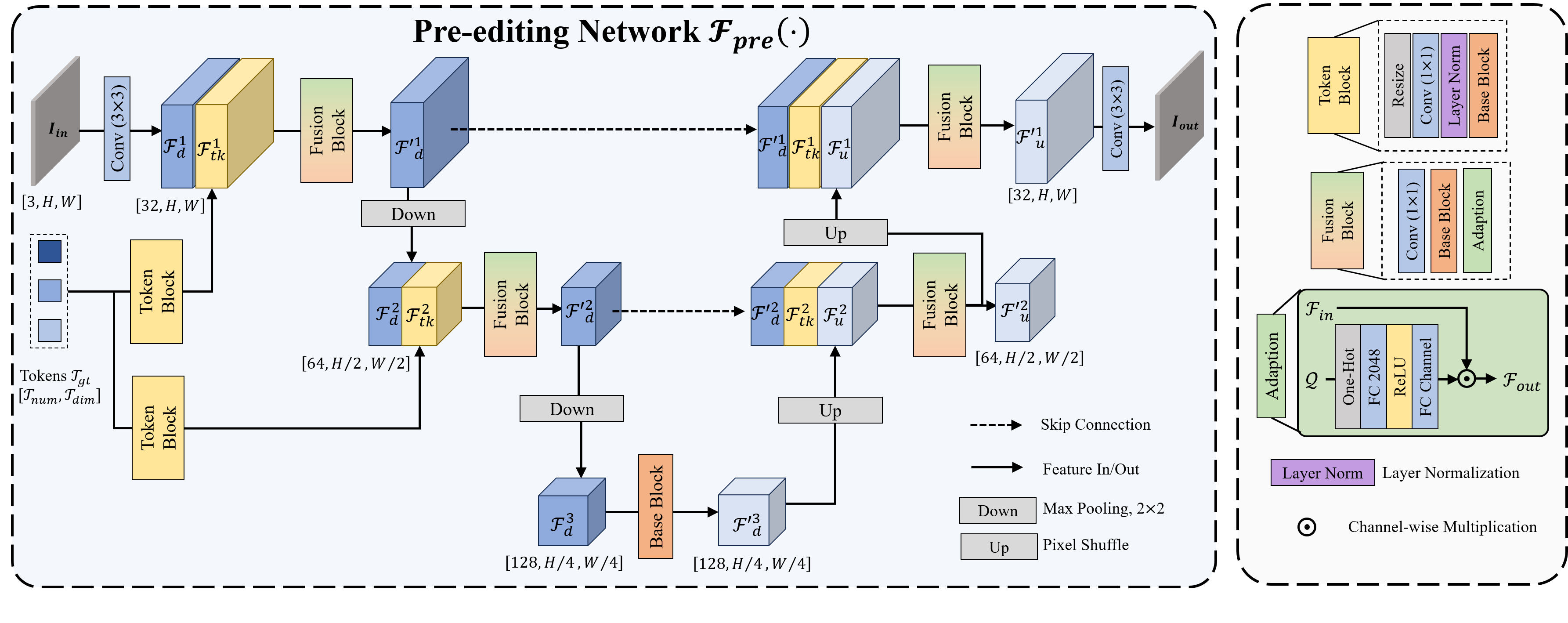}
\caption{ Illustration of the proposed semantic tokens-based pre-editing network. {The proposed pre-editing network consists of three parts: semantic token refinement, down-sampling, and up-sampling. The semantic token refinement branch refines the semantic feature representation in different scales.  The down-sampling and up-sampling branches utilize semantic tokens at multiple scales. }}
\vspace{0.3cm} 
\label{fig_processnet}
\end{figure*}

\subsection{Large Visual-Language Models}

Having witnessed the success of transformer-based models in natural language processing (NLP) tasks, numerous efforts have been made to apply the transformer structure in vision-language tasks.
Parmar \textit{et al.}~\cite{parmar2018image} increased the image size that can be handled by the model by restricting self-attention in local neighborhoods. 
Sparse Transformer~\cite{child2019generating} employs scalable approximations of global self-attention by introducing sparse factorizations of the attention matrix to avoid quadratically growing of the time and memory with the sequence length.
Subsequently, Dosovitskiy \textit{et al.}~\cite{dosovitskiy2020image} discarded the reliance on CNNs and proposed the Vision Transformer (ViT) with excellent results.
To improve the performance with visual modality, efficient pre-training methods~\cite{bao2021beit,he2022masked} were developed.
Based on these advanced results, numerous works~\cite{wang2022ofa,wang2023one,zhou2020unified,wang2021simvlm} achieved superior performance in vision-language tasks by jointly learning multi-modal data.

\section{The Proposed Image Compression Scheme}
\label{section:Framework}

\subsection{Framework in a Nutshell}
\label{section:Nutshell}

This subsection introduces the image compression framework for LVLMs, which is composed of the pre-editing module and the end-to-end codec. 
As shown in Fig.~\ref{fig_paradigm}, the pre-editing module includes the semantic tokens extractor and the pre-editing network, and the end-to-end codec is composed of the encoder and decoder for compressing and reconstructing the preprocessed image. The LVLMs are regarded as the ultimate receivers of the reconstructed images. The text information for different tasks is also fed into LVLMs. 
In detail, the pre-editing module takes the original image $\boldsymbol{I_{o}}$ and compression ratio index $\mathcal{Q}$ as input and outputs the preprocessed image $\boldsymbol{I_{p}}$. The $\mathcal{Q}$ also corresponds to the index of the Lagrange multiplier parameters in the loss function.
Subsequently, the $\boldsymbol{I_{p}}$ is compressed as bitstream $\boldsymbol{\chi} $ by the encoder $Enc\left ( \cdot \right ) $. On the decoder side, the decoder $Dec\left ( \cdot \right ) $ reconstruct the decoded image  $\boldsymbol{I_{d}}$. 
% Finally,  the $\boldsymbol{I_{d}}$ and other modality information are facilitated to different multi-modal tasks with the large models. 
Finally, the large models use the $\boldsymbol{I_{d}}$ and other modality information in different multi-modal tasks. 
In particular, the pre-editing module consists of the tokens exactor $\mathcal{F}_{tk}\left ( \cdot \right ) $ and the pre-editing network $\mathcal{F}_{pre}\left ( \cdot \right ) $.
Given $\boldsymbol{I_{o}}$, the $\mathcal{F}_{tk}\left ( \cdot \right ) $ outputs the tokens $\mathcal{T}_{gt}$ for semantic information extraction. Based on the $\mathcal{T}_{gt}$ and $\boldsymbol{I_{o}}$, the pre-editing network $\mathcal{F}_{pre}\left ( \cdot \right ) $ generates the $\boldsymbol{I_{p}}$ for better rate-accuracy performance.

\begin{figure}[t]
\centering
\includegraphics[width=0.5\textwidth]{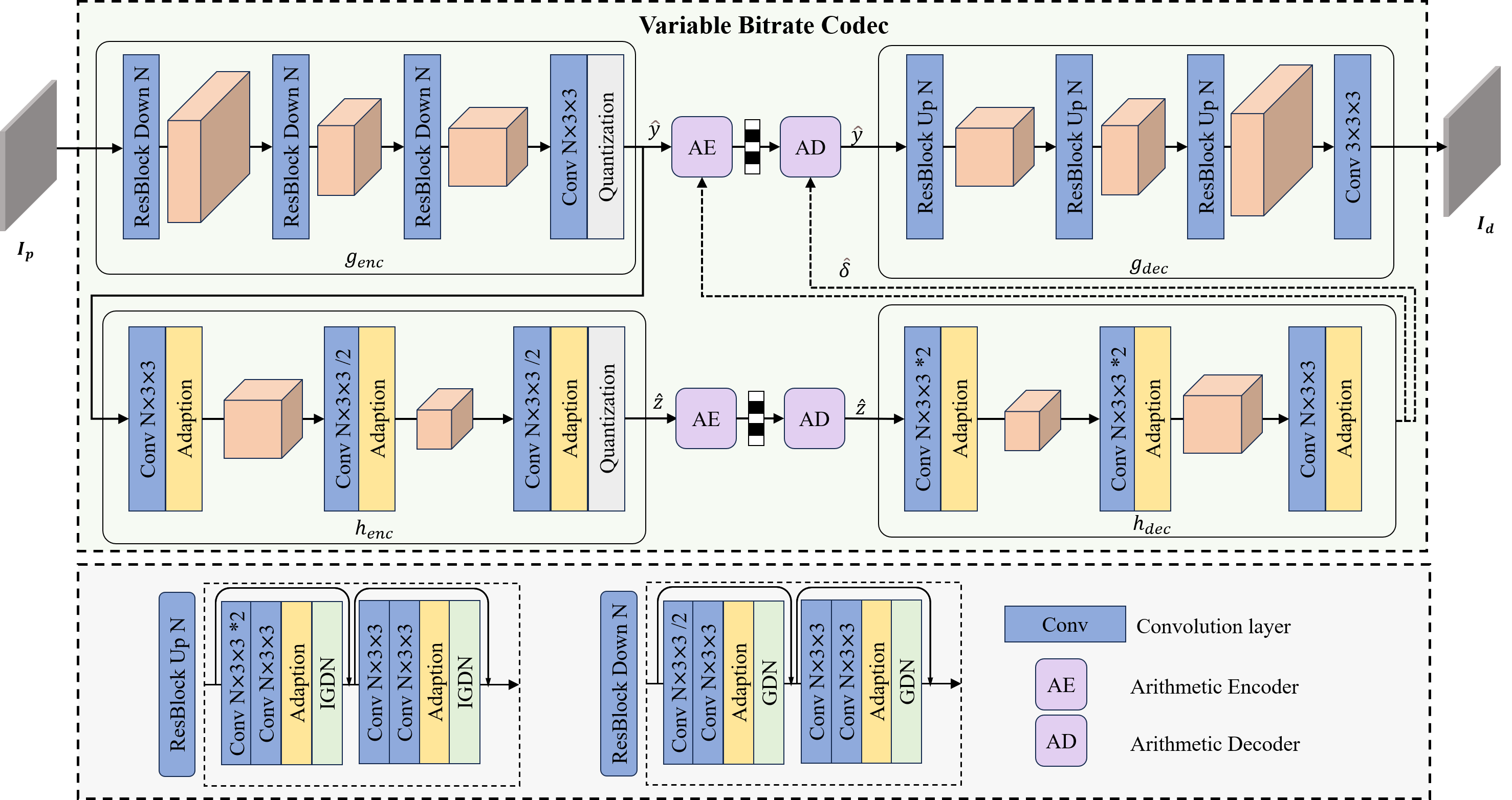}
\caption{ {The paradigm of the proposed variable bitrate codec. The codec includes the $g_{enc}$, $g_{dec}$, $h_{enc}$, and $h_{dec}$, which are composed of the convolutions and compression ratio adaption layers.} }
\vspace{0.3cm} 
\label{fig_CodecNet}
\end{figure}

\subsection{Pre-editing Based on Semantic Tokens}
\label{section:Editing}

In this subsection, we incorporate the semantic tokens into the pre-editing network, accommodating the unique characteristics of LVLMs. Instead of the task-wise optimization in the traditional coding for machine, we formulate the problem with the general semantic information extraction and optimization pipeline, leading to a more robust, effective, and generic solution. 
In particular, given the outstanding performance of visual transformer-based models in various tasks~\cite{li2022exploring,kirillov2023segment}, it is reasonable to assume that the tokens used as internal features could provide accurate and generalized semantics.
To enhance semantic information and remove irrelevant features, we introduce tokens from visual transformers as semantic prompts for the image editing network. 
As shown in Fig.~\ref{fig_processnet}, we combine U-Net~\cite{ronneberger2015u} and token information $\mathcal{T}_{gt}$ to create a pre-editing network. By utilizing semantic tokens at multiple scales, we can achieve better RA performance across a range of tasks.

More specifically, the proposed pre-editing network consists of three parts: semantic token refinement, down-sampling, and up-sampling branches.
The semantic tokens refinement branch uses the token block to extract semantic features $\mathcal{F}_{tk}^{i} $ from the $\mathcal{T}_{gt}$, where the token blocks refine the semantic feature representation in different scales. These features are fed into the down-sampling and up-sampling branches.
The down-sampling branch combines the image features $\mathcal{F}_{d}^{i} $ extracted from the input image and  $\mathcal{F}_{tk}^{i}$, and the fusion block further extracts intermediate features ${\mathcal{F}^{\prime}}_{d}^{i+1}$. Subsequently, ${\mathcal{F}^{\prime}}_{d}^{i}$ is down-sampled to ${\mathcal{F}^{\prime}}_{d}^{i+1}$ by the maximum pooling layer with a $2\times2$ kernel.
${\mathcal{F}^{\prime}}_{d}^{i}$ is fed to the up-sampling branch by skip connections. 
Analogously, the up-sampling branch also combines the intermediate features ${\mathcal{F}}_{u}^{i}$, ${\mathcal{F}}_{tk}^{i}$  as well as ${\mathcal{F}^{\prime}}_{d}^{i}$, and employs the fusion modules to obtain the features ${\mathcal{F}^{\prime}}_{u}^{i}$.
The output image is obtained based on ${\mathcal{F}^{\prime}}_{u}^{i}$ by the convolution layer.

Furthermore, as shown in Fig.~\ref{fig_processnet}, the token block consists of the resize step, the convolution layer with the $1\times1$ kernel, the normalization layer~\cite{ba2016layer}, and the base block. To adapt to images of different scales, the resize step reshapes the tokens from $\left [ \mathcal{T}_{dim}, \mathcal{T}_{num} \right ] $ to $\left [ \mathcal{T}_{dim}, \sqrt{\mathcal{T}_{num}}, \sqrt{\mathcal{T}_{num}} \right ] $ and interpolates the tokens to the size of $\mathcal{F}_{d}^{i} $.
% The base block is the same as the baseline’s block~\cite{chen2022simple} with the channel-wise attention mechanism~\cite{hu2018senet} and improves the performance of the process network.
The base block follows the one in~\cite{chen2022simple} with the channel-wise attention mechanism~\cite{hu2018senet}.
Moreover, the fusion block is composed of the convolution layer for reducing the channel size, the base block for feature enhancement, and the adaption processing.
Given the index $\mathcal{Q}$, the adaption layer predicts the feature weighting vector which is channel-wisely multiplied with the input feature $\mathcal{F}_{in}$ to output modulated feature $\mathcal{F}_{out}$, ultimately adapting the compression ratio.

\subsection{Variable Bitrate Codec}
\label{section:Codec}

This subsection introduces the end-to-end variable bitrate codec that compresses the input image $\boldsymbol{I_{p}}$ and reconstructs the $\boldsymbol{I_{d}}$ with the compression ratio index $\mathcal{Q}$. To remove the redundancy when considering the ultimate receivers are LVLMs, we develop a codec using the variational autoencoders-based codec~\cite{balle2018variational,begaint2020compressai} as the foundation, as shown in Fig.~\ref{fig_CodecNet}. Furthermore, the ResBlcok and compression ratio adaption layers are also introduced to achieve better variable RA performance.
In particular, the codec includes the $g_{enc}$, $g_{dec}$, $h_{enc}$, and $h_{dec}$. 
The $g_{enc}$ and $g_{dec}$, which are composed of convolutions and the ResBlocks, form the image autoencoder structure to compress the image. Meanwhile, the $h_{enc}$ and $h_{dec}$ consist of the convolutions and adaption layers to implement the hyperprior autoencoder for transmitting the spatial distribution of standard deviations $\hat{\delta}$.
Moreover, the Qn is the quantization, and AE and AD represent the arithmetic encoder and arithmetic decoder, respectively. 
Besides, the ResBlocks include convolutions, adapt layers, and GDN/IGDN activation functions. 
For the convolution layer, the parameters N$\times$3$\times$3 represent the number of channels and the kernel size. Herein, $/2$ and $*2$ indicate downsampling or upsampling with the convolution stride and pixel shuffle function, respectively. 
Finally, the adaption layer is consistent with the adaption layer of the pre-editing network.

\begin{figure*}[t]
\centering

% \vspace{0.4cm} 

\begin{minipage}[b]{0.31\linewidth}
  \centerline{\includegraphics[width=\textwidth]{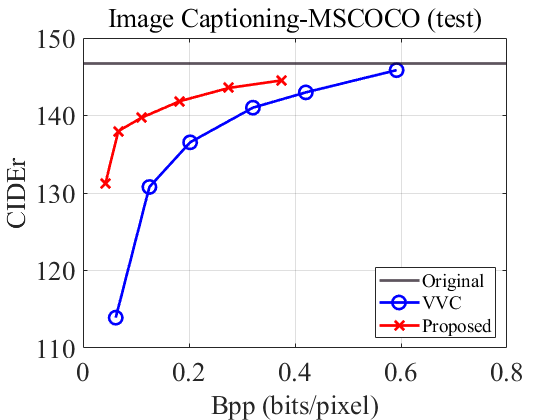}}
  \centerline{ \small (a)}
\end{minipage}
\begin{minipage}[b]{0.31\linewidth}
  \centerline{\includegraphics[width=\textwidth]{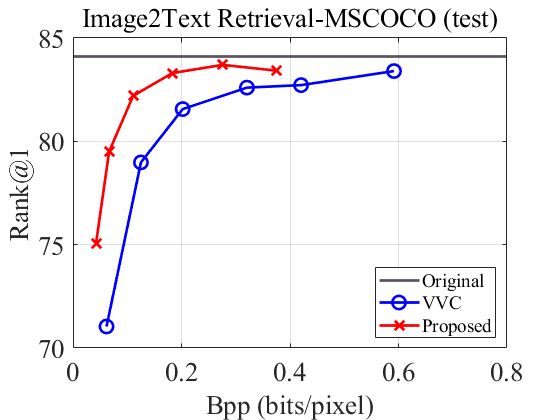}}
  \centerline{ \small (b)}
\end{minipage}
\begin{minipage}[b]{0.31\linewidth}
  \centerline{\includegraphics[width=\textwidth]{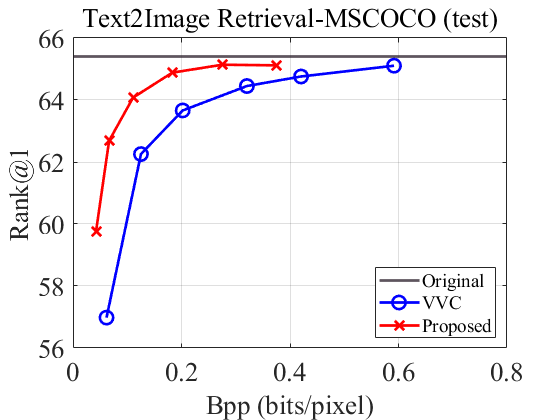}}
  \centerline{ \small  (c)}
\end{minipage}

\vspace{0.3cm} 

\begin{minipage}[b]{0.31\linewidth}
  \centerline{\includegraphics[width=\textwidth]{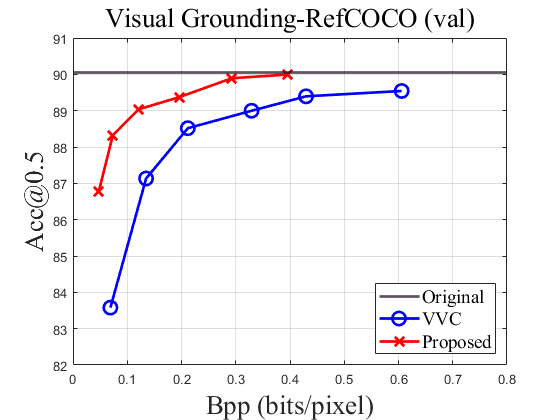}}
  \centerline{ \small (d)}
\end{minipage}
\begin{minipage}[b]{0.31\linewidth}
  \centerline{\includegraphics[width=\textwidth]{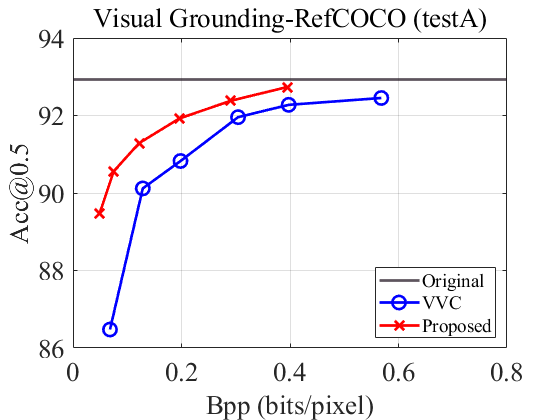}}
  \centerline{ \small (e)}
\end{minipage}
\begin{minipage}[b]{0.31\linewidth}
  \centerline{\includegraphics[width=\textwidth]{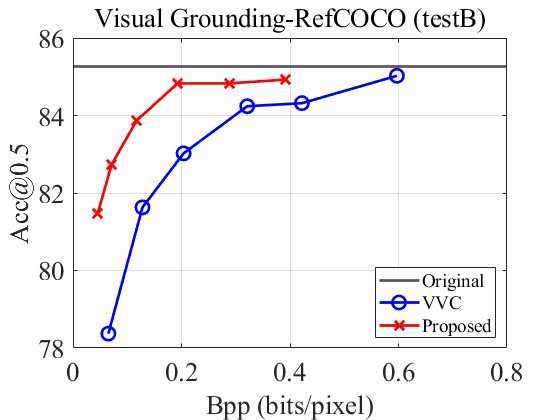}}
  \centerline{ \small  (f)  }
\end{minipage}

\vspace{0.3cm} 

\begin{minipage}[b]{0.31\linewidth}
  \centerline{\includegraphics[width=\textwidth]{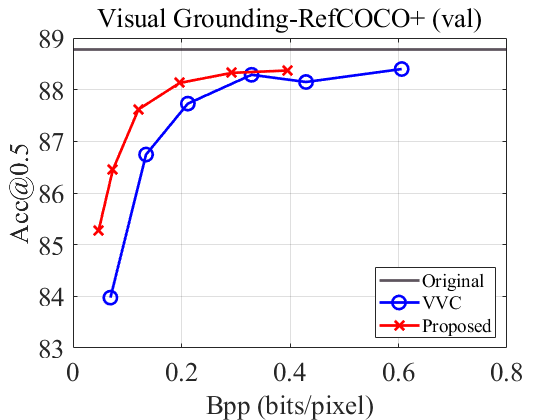}}
  \centerline{ \small (g)}
\end{minipage}
\begin{minipage}[b]{0.31\linewidth}
  \centerline{\includegraphics[width=\textwidth]{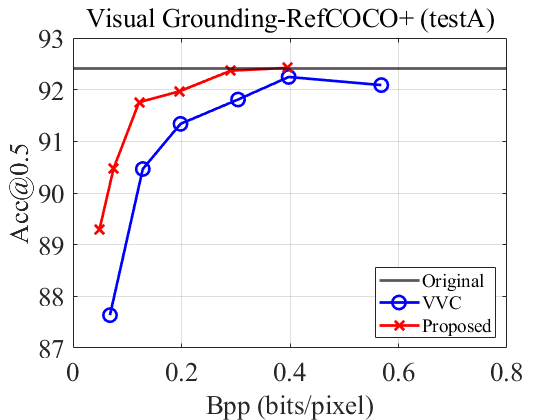}}
  \centerline{ \small (h)}
\end{minipage}
\begin{minipage}[b]{0.31\linewidth}
  \centerline{\includegraphics[width=\textwidth]{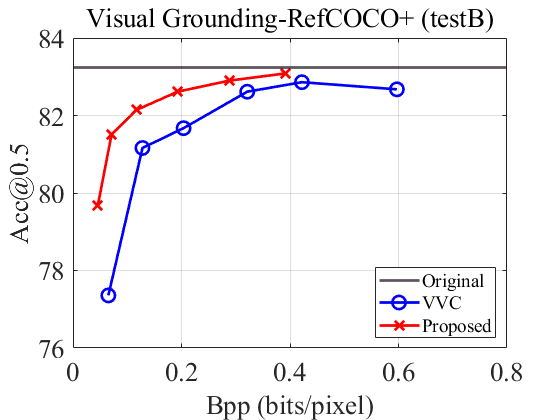}}
  \centerline{ \small  (i)  }
\end{minipage}

\caption{RA performance comparisons with VVC anchor for image captioning, image-text retrieval, and vision grounding tasks that are completed with LVLMs.  }
\label{All_RAresult}

\end{figure*}

\subsection{Loss Functions}
\label{section:Loss}

The training of a traditional codec can be typically casted into a classical rate-distortion optimization (RDO) problem. In general, the training loss function typically consists of the number of coding bits and the distortion between the input and reconstructed signals.
Moreover, to address the problem of coding for machine, the loss from the machine analytics task is also included in the loss function. However, such loss functions cannot be straightforwardly employed in the context of LVLMs, as LVLMs can complete a series of tasks simultaneously. 
By contrast, the loss function of one or several particular tasks focuses only on task-related semantics, ignoring irrelevant semantic information that might be meaningful for other tasks. Therefore, task-based loss functions in training make it challenging to encompass the semantics of the entire image comprehensively. 
To address this challenge, the loss functions based on tokens instead of tasks are incorporated to improve the generalization capability and RA performance of the LVLMs. 
Consequently, the loss function of the proposed framework is formulated as follows,
\begin{equation}
\label{loss_all}
    \mathcal {L}_{} = \lambda_{R} \mathcal {L}_{R} + \lambda_{D}\mathcal {L}_{D} + \lambda_{T}\mathcal {L}_{T}, 
\end{equation}
where $\mathcal {L}_{R}$, $\mathcal {L}_{D}$, and $\mathcal {L}_{T}$ are the coding bitrate loss, image distortion loss, and token-based loss, respectively. Meanwhile, the $\lambda_{R}$, $\lambda_{D}$, and $\lambda_{T}$ are the hyper-parameters of the weights for the corresponding loss. The $\mathcal {L}_{R}$ is the bitrate loss obtained by the proposed codec. 
The image distortion loss $\mathcal {L}_{D}$ is the MSE loss between the $\boldsymbol{I_{o}}$ and $\boldsymbol{I_{d}}$.

More specifically, the token-based loss includes the mean squared error (MSE) loss $\mathcal {L}_{tk}$ and the rank loss $\mathcal {L}_{rk}$. 
\begin{equation}
\mathcal {L}_{T} = \lambda_{tk}\mathcal {L}_{tk} + \lambda_{rk}\mathcal {L}_{rk},
\end{equation}
The $\mathcal {L}_{tk}$ is also the MSE loss between the ground truth tokens $\mathcal{T}_{gt}$ and tokens $\mathcal{T}_{d}$ of reconstructed image $\boldsymbol{I_{d}}$. The MSE loss of tokens facilitates the reduction of compression distortion at the semantic level. Moreover, the feature collapse phenomenon implies the relevance of the rank of the tokens to the semantic information~\cite{dong2021attention,wang2023pangu}. The larger the rank of the tokens, the more likely that more semantics are included. 
As such, the rank loss is proposed to measure the loss of the rank of the tokens during compression and is used to optimize information loss at the semantics level. The rank of the matrix is discontinuous, such that we use the trace instead of the rank.
In detail, the rank loss $\mathcal {L}_{rk}$ measures MSE loss between the rank of the ground truth tokens and the rank of reconstructed tokens, which is formulated as,
\begin{equation}
\label{}
\mathcal {L}_{rk} =  D ( \mathrm{tr} (\mathrm{Sig}(\Sigma _{gt} )) , \mathrm{tr}(\mathrm{Sig}(\Sigma _{d} )) ) ,
\end{equation}
where the $\Sigma _{gt}$ and $\Sigma _{d}$ are the eigenvalue matrix of $\mathcal{T}_{gt}$ and $\mathcal{T}_{d}$, respectively. The $\mathrm{Sig}(\cdot)$ and $\mathrm{tr}(\cdot)$ mean the sigmoid function and the trace of the input matrix, respectively. {The function $D(\cdot, \cdot)$ defines the MSE loss between the rank of ground truth tokens and the reconstructed tokens. }

\begin{table*}[t]
\caption{ Experimental results of the proposed framework for the vision-language tasks with two LVLMs.} 
\begin{center}
\begin{tabular}{ccccc}
% \begin{tabular}{m{30pt}<{\centering} m{50pt}<{\centering} m{30pt}<{\centering} m{40pt}<{\centering} m{40pt}<{\centering}}
\hline
Number & Machine Task                       & Task Model  & Dataset (Splits)           & BD-Rate (Anchor: VVC~\cite{VVC})            \\ \hline
1               & Image Captioning                   & OFA~\cite{wang2022ofa} & MSCOCO (test)              &  -63.25\% \\
2               & Image to Text Retrieval            & OP~\cite{wang2023one}  & MSCOCO (test)              &  -53.02\% \\
3               & Text to Image   Retrieval          & OP~\cite{wang2023one} & MSCOCO (test)              &  -53.29\% \\
4               & \multirow{2}{*}{Visual Grounding} & OFA~\cite{wang2022ofa} & RefCOCO (val/testA/testB)  &  -62.19\% / -48.21\%/ -61.38\% \\
% 5               &                                    &OP~\cite{wang2023one} & RefCOCO (val/testA/testB)  &  -53.14\%  /	-57.65\% / --61.22\% \\
% 6               &                                    & OFA~\cite{wang2022ofa} & RefCOCO+ (val/testA/testB) &  -54.14\% / -54.10\% / -59.34\% \\
5               &                                    & OP~\cite{wang2023one} & RefCOCO+ (val/testA/testB) &  -38.76\% /	-46.29\% / -51.31\% \\ 
\hline
\end{tabular}
\end{center}
% \vspace{0.3cm} 
\label{AllPerformance}
\end{table*}

\begin{table*}[t]
\caption{ Performance comparisons for the image captioning and image-text retrieval tasks based on LVLMs in ablation studies (anchor: VVC).} 
\begin{center}
\begin{tabular}{cccc}
\hline
\multirow{2}{*}{Methods} & \multicolumn{3}{c}{Machine task}                                     \\ \cline{2-4} 
                         & Image Captioning & Image to Text Retrieval & Text to Image Retrieval \\ \hline
Proposed                 & -63.25\%           & -53.02\%                  & -53.29\%                  \\
Proposed (w/o Codec Training) & -29.09\%           & -21.40\%                  & -23.30\%                 \\
Proposed (w/o Pre-editing)            & -49.77\%           & -36.87\%                  & -40.71\%                 \\
Proposed (w/o $\mathcal {L}_{rk}$)  & -61.24\%           & -50.20\%                  & -49.96\%                \\ 
{Proposed (w/o Tokens)}  & -57.03\%           & -47.84\%                  & -46.73\%                \\ \hline
\end{tabular}
\end{center}
% \vspace{0.3cm} 
\label{AblationTable}
\end{table*}

\begin{figure*}[t]
\centering

\begin{minipage}[b]{0.31\linewidth}
  \centerline{\includegraphics[width=\textwidth]{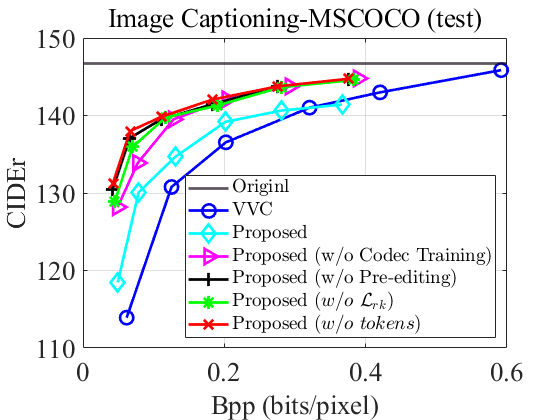}}
  \centerline{ \small (a)}
\end{minipage}
\begin{minipage}[b]{0.31\linewidth}
  \centerline{\includegraphics[width=\textwidth]{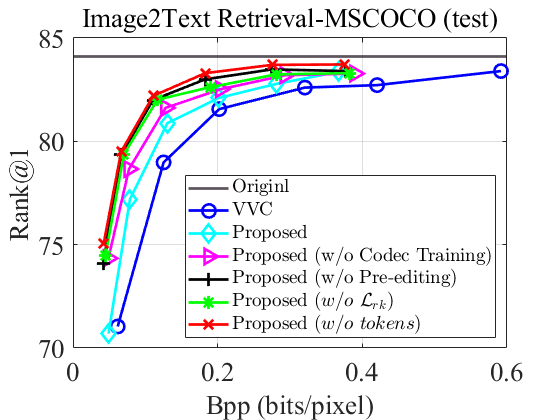}}
  \centerline{ \small (b)}
\end{minipage}
\begin{minipage}[b]{0.31\linewidth}
  \centerline{\includegraphics[width=\textwidth]{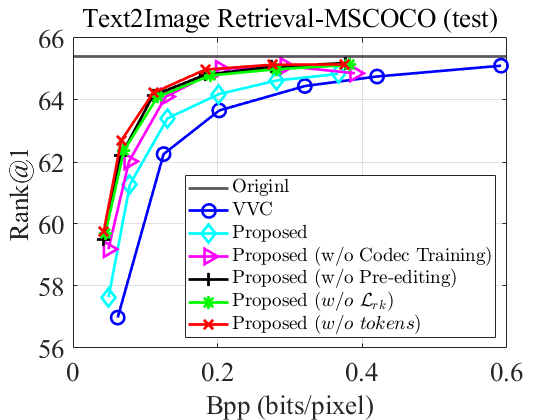}}
  \centerline{ \small  (c)  }
\end{minipage}

% \vspace{0.3cm} 

\caption{Ablation studies for image captioning and image-text retrieval tasks based on LVLMs.  }

\label{Fig_AlbStudy}

\end{figure*}

\begin{figure*}[t]
\centering
\includegraphics[width=0.85\textwidth]{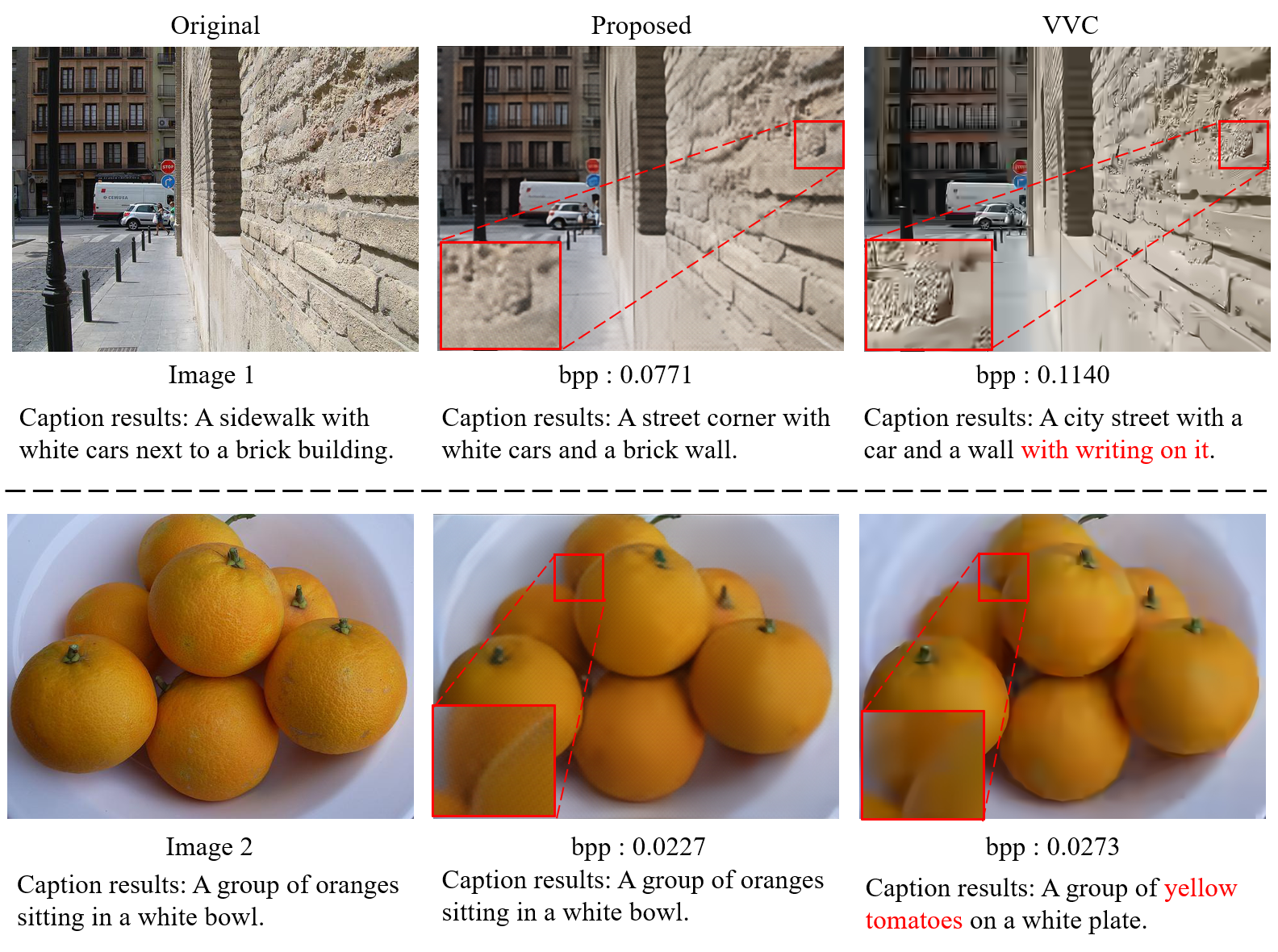}
\caption{ {Illustration of the reconstruction images and image caption results for VVC and the proposed method. The image caption is based on the OFA multimodal large model~\cite{wang2022ofa}.
% For each input image, the top row shows the reconstruction image, the next row shows the bpp, and the last row shows the image caption based on the OFA model~\cite{wang2022ofa}. 
The first column shows the original images and the ground truth of image captions. The second and third columns show the images reconstructed by the proposed method and baseline method, respectively. In the image caption, the red text is the error information compared with the ground truth.} }
\vspace{0.3cm} 
\label{fig_visual}
\end{figure*}

\section{Experimental Results}
\label{section:Experiments}

\subsection{Experimental Setup}

\subsubsection{Datasets and Task Models}  
The LVLMs OFA~\cite{wang2022ofa} and OP~\cite{wang2023one} are treated as the ultimate consumers of the decoded images. 
In training, the COCO dataset~\cite{lin2014microsoft} is used to train the proposed variable bitrate framework, and semantic tokens in training and testing are extracted by the backbone of the ViTDet~\cite{li2022exploring}.
To evaluate the influence of compression on cross-modal downstream tasks, the MSCOCO dataset~\cite{lin2014microsoft,young2014image,chen2015microsoft} and RefCOCO/RefCOCO+~\cite{kazemzadeh2014referitgame,yu2016modeling} are used as testing datasets.
For both the RefCOCO and RefCOCO+ datasets, val/testA/testB are regarded as different testing splits. 
% For both the RefCOCO datasets, val/testA/testB are regarded as different test splits. 
In testing, the OFA~\cite{wang2022ofa} is taken as the task network for the visual grounding task and image captioning task, and OP~\cite{wang2023one} is used for the visual grounding task and image-text retrieval task.
Moreover, we use the Acc@0.5, R@1, and CIDEr~\cite{vedantam2015cider} as the evaluation metrics on visual grounding, image-text retrieval, and image captioning tasks, respectively.
Furthermore, the Bjontegaard Delta Bit Rate (BD-Rate)~\cite{bjontegaard2001calculation} is used to measure the percentage of saved bitrate at the same task accuracy. 
In addition, for the nonmonotonic curves, the pareto front curve is created to present the performance~\cite{pareto}.

\subsubsection{Implementation Details} 
The Adaptive Moment Estimation (Adam)~\cite{Adam} optimizer with the parameters ($\beta_1 = 0.5$ and $\beta_2=0.999$) is set as the optimizer. The model training is conducted on the NVIDIA TESLA A100 GPU with a batch size of 8.
To achieve better performance, we divide the training process of the scheme into three stages.  In the first stage, the variable bitrate codec is trained without token-related loss, which is taken as the pre-trained codec. For this stage, the initial learning rate is set to $1 \times 10^{-4} $ with 200K iterations. 
In the second stage, the pre-editing network is trained with the pre-trained codec, and the initial learning rate is set to $1 \times 10^{-4} $ and reduced to $1 \times 10^{-6} $  with the cosine annealing schedule~\cite{loshchilov2016sgdr} in 150K iterations.
After that, the pre-trained codec and pre-editing network are fine-tuned in the third stage.
Herein, the initial learning rate is set as $1 \times 10^{-5} $ to the codec and $1 \times 10^{-6} $ to the pre-editing network with 150K iterations.

\subsection{Performance Evaluations}

In this subsection, we conduct a series of experiments to study the generalization capability and efficiency of the proposed framework.
First, the RA performance of the proposed framework is compared with the baseline across different vision-language tasks and corresponding LVLMs.
Subsequently, we conduct the ablation studies to investigate the effect of different modules in the proposed framework. 

\subsubsection{Overall Performance Comparisons}

To verify the efficiency and the generalization capability of the proposed image compression framework, the RA performance of the proposed method and VVC anchor are compared in different vision-language tasks and LVLMs.
For the VVC anchor, the VTM-22.2~\cite{VTMSoft} is deployed as the VVC platform with coding performed in the YUV420 format in these experiments, which is the state-of-the-art coding standard. The QPs in testing are set as $\left [  32, 35, 37, 40, 43, 47 \right ] $  to achieve a similar bits-per-pixel (bpp) range with the proposed framework. In addition, the BD-Rate is used for performance evaluations.

As shown in Table~\ref{AllPerformance}, the proposed framework performs better than the anchor in these experiments.
Moreover, in Fig.~\ref{All_RAresult}, we compare RA curves for four different tasks with two large models, respectively. 
We can observe that the proposed framework achieves promising improvements compared to the anchor method across different datasets and tasks.  
Meanwhile, these proposed curves are close to performance obtained by the analysis of the original images. 
These experimental results demonstrate the proposed framework surpasses the anchor method, maintaining the performance of multi-modal tasks while significantly saving bit rates.
{Furthermore,  Fig.~\ref{fig_visual} shows two examples of reconstructed images and image captioning results. It is obvious that the proposed compression framework could achieve accurate image captioning results with less bitrate consumption than VVC. Meanwhile, we enlarge the patch of the decoded image to show the details. As shown in Image 1 of Fig.~\ref{fig_visual}, the proposed method does not introduce any undesired textures. In Image 2 of Fig.~\ref{fig_visual}, the proposed method results in clearer object boundaries compared to the baseline.} In conclusion, these experiments demonstrate that the proposed framework has a promising generalization capability and efficiency toward different vision-language tasks and datasets.

\subsubsection{Ablation Studies}
Herein, the ablation experiments are conducted in three multi-modal tasks to demonstrate the influences of different modules. 
{ In particular, we first adopt the pre-trained codec instead of the codec fine-tuned with token-related loss, denoted as the Proposed (w/o Codec Training).
Second, we remove the model of the pre-editing such that the given images are coded directly, denoted as the Proposed (w/o Pre-editing). 
Moreover, the proposed framework is retrained and evaluated without the rank loss, denoted as Proposed (w/o Rank Loss), to study the performance of the proposed rank loss.
Finally, the pre-editing module of the proposed framework removes the tokens inputs, denoted as Proposed (w/o Tokens), to evaluate the performance of semantic information guidance.} 
Based on the results in Table~\ref{AblationTable} and Fig.~\ref{Fig_AlbStudy}, we can draw some conclusions.
Firstly, the proposed method achieves the best performance, which demonstrates the rationality and effectiveness of this design. 
Compared with the Proposed (w/o $\mathcal {L}_{rk}$), the proposed scheme achieves 2.01\%, 2.82\%, and 3.33\% BD-Rate savings in three tasks, respectively, illustrating the effectiveness of rank loss.
Finally, the Proposed (w/o Pre-editing) achieves better performance than the Proposed (w/o Codec Training), which may be because the codec can adapt to the feature distribution rather than just enhancing features and eliminating redundancy.

\begin{table}[t]
\caption{ FLOPS, number of parameters and inference time of different modules.} 
% \vspace{0.3cm} 
\begin{center}
\begin{tabular}{m{70pt}<{\centering} m{40pt}<{\centering} m{50pt}<{\centering} m{40pt}<{\centering}}
\hline
Networks  & FLOPS (G) & Number of Params (M) & Inference Time (s) \\ \hline
Tokens Extractor Network & 365.8    & 85.64  & 0.0436   \\ \hline
Pre-editing Network  & 2.420    & 23.51  & 0.0248   \\ \hline
Encoder Network & 12.22   & 30.97     & 0.0215 \\ \hline
Decoder Network & 12.11   & 30.45   & 0.0203  \\ \hline
\end{tabular}
\end{center}

\label{Complexity}
\end{table}

\subsubsection{Complexity Study}
When deploying compression frameworks in real-world scenarios, computational complexity plays a critical role. Therefore, complexity study experiments are conducted to measure the number of parameters, inference time, and floating-point operations per second (FLOPS) of four networks in the framework.
Experimental results are shown in Table~\ref{Complexity} with 256$\times$256 inputs.  The testing condition is NVIDIA Tesla V100 GPU with 32GB memory. 
In particular, the decoder network's inference time is 0.0203s, suggesting that the proposed framework shows the real-time decoding potential in real-world applications. 

\section{Conclusion}
\label{section:Conclusion}

In this paper, we have proposed a new solution for image compression when LVLMs are treated as visual data consumers. The novelty of this paper lies in that the learning process is fully guided by semantically meaningful tokens, which enables the reconstructed images to preserve semantic information in a limited bandwidth environment. Moreover, a dedicatedly designed pre-processing model is proposed for RA performance improvement by discarding the irrelevant semantic information in images.
The proposed scheme could adapt to diverse multi-modal tasks with LVLMs. The framework proposed in the experiment achieves superior RA performance compared to the state-of-the-art coding standard. Meanwhile, the experiments on different multi-modal tasks and datasets verify the generalization ability of this framework, demonstrating its potential in practical applications. The proposed framework validates the potential of video coding for large models and provides a reference for the compression research of multi-modal data for large models. In the future, the proposed scheme can be expanded to multi-modal data, various scenarios, and diverse large model tasks. Meanwhile, the non-monotonous RA curve also indicates the requirement for research on the robustness of a large model in real-world applications.

\appendices

\ifCLASSOPTIONcaptionsoff
  \newpage
\fi

\bibliographystyle{IEEEtran}
\bibliography{refs}

\end{document}